\documentclass[11pt,a4paper]{article}
\usepackage[hyperref]{AACL-IJCNLP2020}
\usepackage{times}
\usepackage{latexsym}

\usepackage{multirow}

\usepackage{microtype}

\aclfinalcopy % Uncomment this line for the final submission
\def\aclpaperid{406} %  Enter the aacl-ijcnlp Paper ID here

\usepackage{hyperref}

\usepackage{subcaption}
\usepackage{graphicx, booktabs}

\title{Toxic Language Detection in Social Media for Brazilian Portuguese:\\ New Dataset and Multilingual Analysis}

\author{João A. Leite, Diego F. Silva \\
  Departamento de Computação \\
  Federal University of São Carlos \\
  São Carlos, Brazil \\
  \texttt{joaoaugustobr@hotmail.com} \\
  \texttt{diegofs@ufscar.br} \\ \And
  Kalina Bontcheva, Carolina Scarton \\
  Department of Computer Science \\
  University of Sheffield \\
  Sheffield, UK \\
  \texttt{k.bontcheva@sheffield.ac.uk} \\
  \texttt{c.scarton@sheffield.ac.uk}\\}

\date{}

\begin{document}
\maketitle
\begin{abstract}
Hate speech and toxic comments are a common concern of social media platform users. Although these comments are, fortunately, the minority in these platforms, they are still capable of causing harm. Therefore, identifying these comments is an important task for studying and preventing the proliferation of toxicity in social media. Previous work in automatically detecting toxic comments focus mainly in English, with very few work in languages like Brazilian Portuguese. In this paper, we propose a new large-scale dataset for Brazilian Portuguese with tweets annotated as either toxic or non-toxic or in different types of toxicity. We present our dataset collection and annotation process, where we aimed to select candidates covering multiple demographic groups. State-of-the-art BERT models were able to achieve 76\% macro-$F1$ score using monolingual data in the binary case. We also show that large-scale monolingual data is still needed to create more accurate models, despite recent advances in multilingual approaches. An error analysis and experiments with multi-label classification show the difficulty of classifying certain types of toxic comments that appear less frequently in our data and highlights the need to develop models that are aware of different categories of toxicity. 
\end{abstract}

\section{Introduction}
Social media can be a powerful tool that enables virtual human interactions, connecting people and enhancing businesses' presence. On the other hand, since users feel somehow protected under their virtual identities, social media has also become a platform for hate speech and use of toxic language. Although hate speech is a crime in most countries, identifying cases in social media is not an easy task, given the massive amounts of data posted every day. Therefore, automatic approaches for detecting online hate speech have received significant attention in recent years \cite{waseem-hovy-2016-hateful,davidson-etal-2017,zampieri-etal-2019-semeval}.
In this paper, we focus on the analysis and automatic detection of \textbf{toxic comments}. Our definition of toxic is similar to the one used by the Jigsaw competition,\footnote{\url{https://www.kaggle.com/c/jigsaw-toxic-comment-classification-challenge/overview}} where comments containing insults and obscene language are also considered, besides hate speech.\footnote{This is also similar to the usage of \textit{offensive} comments in OffensEval \cite{zampieri-etal-2019-semeval,zampieri-etal-2020-semeval}.} Systems capable of automatically identifying toxic comments are useful for platform's moderators and to select content for specific users (e.g. children).
Nevertheless, there are multiple challenges specific to process toxic comments automatically, e.g. (i) toxic language may not be explicit, i.e. may not contain explicit toxic terms; (ii) there is a large spectrum of types of toxicity (e.g. sexism, racism, insult); (iii) toxic comments correspond to a minority of comments, which is fortunate, but means that automatic data-driven approaches need to deal with highly unbalanced data. 

Although there is some work on this topic for other languages -- e.g. Arabic \cite{mubarak-etal-2017-abusive} and German \cite{wiegand-etal-2018-germeval} --, most of the resources and studies available are for English \cite{davidson-etal-2017,wulzyn-etal-2017,founta-etal-2018,mandl-etal-2019,zampieri-etal-2019-semeval}.\footnote{A large list of resources is available at \url{http://hatespeechdata.com}.} 
For Portuguese, only two previous works are available \cite{fortuna-etal-2019-hierarchically,depelle-2017} and their datasets are considerably small, mainly when compared to resources available for English. %\cite{fortuna-etal-2019-hierarchically}'s dataset represents an important advance, however, its size and complex hierarchy are limiting factors for training reliable deep learning models. 
%Despite \newcite{fortuna-etal-2019-hierarchically} representing an important advance, their dataset is still small and they address specifically hate speech detection, while our focus is on the broader  concept of toxic comments. %Despite the small amount of previous work in Portuguese, Brazil is ranked sixth regarding the number users in Twitter (approximately $14.35$ million users).\footnote{\url{http://www.statista.com/statistics/}} Also, Brazilian society is highly polarised, which increases the number of offensive interactions and justify the need of more research for this language.

We present \textbf{ToLD-Br} (Toxic Language Dataset for Brazilian Portuguese), a new dataset with Twitter posts in the Brazilian Portuguese language.\footnote{It is important to distinguish the language variant, since there are multiple differences between Brazilian Portuguese lexicon and other variants of Portuguese.} %We collected a total of $10$ million tweets and 
A total of $21$K tweets were manually annotated into seven categories: \textit{non-toxic}, \textit{LGBTQ+phobia}, \textit{obscene}, \textit{insult}, \textit{racism}, \textit{misogyny} and \textit{xenophobia}. %Tweets were manually annotated %by volunteers from a university in Brazil, 
Each tweet has three annotations that were made by volunteers from a university in Brazil. Volunteers were selected taking into account demographic information, aiming to create a dataset as balanced as possible in regarding to demographic group biases.  
This is then the largest dataset available for toxic data analysis in social media for the Portuguese language and the first dataset with demographic information about annotators.\footnote{ToLD-Br is available at: \url{https://github.com/JAugusto97/ToLD-Br}}%\footnote{The full dataset, including information about the splits used in our experiments, will be made available in the paper's final version under a CC BY-SA 4.0 license (a sample is available as Supplementary Material).} %An in-depth analysis of this dataset shows aggregated information about the different classes and different demographic groups.

We experiment with Brazilian Portuguese~\cite{souza2019portuguese} and Multilingual~\cite{Wolf2019HuggingFacesTS} BERT models \cite{devlin-etal-2019-bert} for the binary task of \textbf{automatically classifying toxic comments}, since similar models achieve state-of-the-art results for the same task in other languages \cite{zampieri-etal-2019-semeval}. Models fine-tuned on monolingual data achieve up to $76$\% of macro-$F1$, improving $3$ points over a baseline. 
Besides, BERT-based approaches with multilingual pre-trained models enable transfer learning and zero-shot learning. The OffensEval 2019 OLID dataset \cite{zampieri-etal-2019-predicting} is then used to experiment with (i) \textbf{transfer-learning}: where both OLID and ToLD-Br are used to fine-tune BERT; and, (ii) \textbf{zero-shot learning}: where BERT is fine-tuned using only OLID. Results highlight the importance of language-specific datasets, since transfer learning does not improve over monolingual models and zero-shot learning achieves only a macro-$F1$ of $56$\%. %These results .% for this task.

\textbf{An error analysis} is performed using our best model, where the worst-case scenario, i.e., classifying \textit{toxic} comments as \textit{non-toxic}, is further investigated, taking into account the fine-grained categories. Results show that categories with fewer examples in the dataset (\textit{racism} and \textit{xenophobia}) are more likely to be mislabelled than other classes, with the best performance being achieved by majority classes (\textit{insult} and \textit{obscene}). 
We also analyse the \textbf{amount of data} needed in order to achieve the best performance in binary classification. % and find out that over $6$K data points is needed to reach stable precision and recall of around $70$\%.
Models trained with few examples are only accurate in predicting the majority class (\textit{non-toxic}). As the number of instances grow, the performance on the minority class (\textit{toxic}) improves significantly.  
Finally, we experiment with \textbf{multi-label classification}, where each different type of toxicity is automatically classified. This is a considerably harder problem than binary classification, where BERT-based models do not outperform the baseline.

Section \ref{sec:related} presents an overview of relevant previous work. Section \ref{sec:dataset} shows details about the ToLD-Br dataset. Material and methods are presented in Section \ref{sec:materials}, whilst results are discussed in Section \ref{sec:results}. Finally, Section \ref{sec:conclusion} shows a final discussion and future work.

\section{Related Work} \label{sec:related}
Although multiple researchers have addressed the topic of hate speech (e.g. \newcite{waseem-hovy-2016-hateful}, \newcite{chung-etal-2019-conan}, \newcite{basile-etal-2019-semeval}), we focus the literature review on previous work related to toxic comments detection, the topic of our paper. 
Due to space constraints, we only describe papers that create and use Twitter-based datasets and/or focus on the Brazilian Portuguese language.

\paragraph{English} \newcite{davidson-etal-2017} present a dataset with around $25$K tweets annotated by crowd-workers as containing \textit{hate}, \textit{offensive language}, or \textit{neither}. They build a feature-based classifier with TF-IDF transformation over $n$-grams, part-of-speech information, sentiment analysis, network information (e.g., number of replies), among other features. Their best model, trained using logistic regression, achieves a macro-$F1$ of $90$. 
%\cite{wulzyn-etal-2017} 
\newcite{founta-etal-2018} also rely on crowd-workers to annotate $80$K tweets into eight categories: \textit{offensive}, \textit{abusive}, \textit{hateful speech}, \textit{aggressive}, \textit{cyberbullying}, \textit{spam}, and \textit{normal}. They perform an exploratory approach to identify the categories that cause most confusion to crowd-workers. Their final, large-scale annotation is done using four categories: \textit{abusive}, \textit{hateful}, \textit{normal}, or \textit{spam}. OffensEval is a series of shared tasks focusing on offensive comments detection \cite{zampieri-etal-2019-semeval,zampieri-etal-2020-semeval}. The OLID dataset (used in the 2019 edition) has around $14$K tweets in English manually annotated as \textit{offensive} or \textit{non-offensive}. The best model for the relevant task A (\textit{offensive} versus \textit{non-offensive}) uses a BERT-based classifier and achieves $82.9$ of macro-$F1$.

\paragraph{German} A shared task (organized as part of GermEval 2018) aimed to classify tweets in German categorized into \textit{offensive} or \textit{non-offensive} \cite{wiegand-etal-2018-germeval}. They make available a manually annotated dataset with approximately $8.5$K tweets. The best system achieved $76.77$ of $F1$-score and was a feature-based ensemble approach.

\paragraph{Arabic} \newcite{mubarak-etal-2017-abusive} present a dataset with $1.1$K manually annotated tweets into \textit{obscene}, \textit{offensive}, or \textit{clean}. They experiment with lexical-based approaches that achieve a maximum of $60$ $F1$-score. \newcite{mulki-etal-2019-l} create a dataset with tweets in the Levantine dialect of Arabic manually annotated into \textit{normal}, \textit{abusive}, or \textit{hate} (with approximately $5$K tweets). The authors use feature-based approaches to induce models for ternary and binary scenarios, with best systems achieving $74.4$ and $89.6$ of $F1$-score, respectively. 

\paragraph{Spanish} \newcite{alvarez-carmona-etal-2018} present a shared task aiming to detect aggressive tweets in Mexican Spanish. They manually annotate $11$K tweets into \textit{aggressive} or \textit{non-aggressive}. The best system is a feature-based approach with macro-$F1$ of $62$.

\paragraph{Hindi} \newcite{mathur-etal-2018-detecting} present a dataset of around $3.6$K tweets in Hinglish (spoken Hindi written using the Roman script). The dataset was annotated into three classes \textit{not offensive}, \textit{abusive} and \textit{hate-inducing} by ten NLP researchers. A Convolutional Neural Network (CNN) architecture with transfer learning is used, where the model is trained with both Hinglish and English data (from \cite{davidson-etal-2017}), achieving $71.4$\% of $F1$-score.

\paragraph{Portuguese} \newcite{depelle-2017} make available a dataset with $1,250$ comments, extracted from comment sessions of \url{g1.globo.com} website, and annotated them into categories of \textit{offensive} or \textit{non-offensive}. The offensive class was also subdivided into \textit{racism}, \textit{sexism}, \textit{LGBTQ+phobia}, \textit{xenophobia}, \textit{religious in-tolerance}, or \textit{cursing}. They experiment with binary classification, using $n$-grams as features to SVM and NaiveBayes models. Best results are achieved with SVM reaching a weighted F1 score between $77$ and $82$, depending on different label interpretations. \newcite{fortuna-etal-2019-hierarchically} describe a dataset with $5,668$ tweets classified as \textit{hate} vs. \textit{non-hate}, with the \textit{hate} class further classified following a fine-grained hierarchy. Experiments with binary classification show a $F1$ score of $78$ using an LSTM-based architecture.

\paragraph{Multilingual} HASOC was a shared task aiming to classify hate speech and offensive comments in English, German, and Hindi \cite{mandl-etal-2019}. Their dataset contains around $7$K tweets and Facebook posts manually annotated. Sub-task A separates posts into \textit{hate speech} or \textit{offensive} versus \textit{neither}; and, sub-task B separates posts containing \textit{hate speech} or \textit{offence} into three categories: \textit{hate speech}, \textit{offensive} or \textit{profane}. Best performing systems in all languages used deep learning approaches.
For OffensEval 2020 \cite{zampieri-etal-2020-semeval}, a more extensive training data is available for English (over $9$M tweets), although the annotation was made semi-automatically. Arabic, Danish, Greek, and Turkish datasets are also available with manually annotated labels. For all languages, best models are achieved using some variation of BERT.    

Our work is different from previous approaches because we (i) release a large-scale dataset for a language other than English, that was created with the aim to reduce demographic biases; (ii) experiment with multilingual approaches, including transfer learning and zero-shot-learning; (iii) perform an analysis of the amount of data needed to train reliable models; and, (iv) experiment with multi-label classification, providing first insights into this challenge task.
%\url{http://hatespeechdata.com}
%\url{https://docs.google.com/document/d/1WVkVGp29Jt6d-4fBnZ5OWVYuFn_03rzz-KBqPsu6gTM/edit}

\section{Dataset} \label{sec:dataset}
%Diego + João
In this section, we describe the procedure adopted to create ToLD-Br and present its main features.

\subsection{Data collection}
We used the GATE Cloud's Twitter Collector\footnote{\url{https://cloud.gate.ac.uk/shopfront/displayItem/twitter-collector}} to collect posts on the Twitter platform from July to August 2019. We used two different strategies to select tweets for ToLD-Br, aiming to increase the probability of obtaining posts with toxic content, given that the volume of toxic tweets is significantly smaller than data without offensive language.
Our first strategy searches for tweets that mention predefined hashtags or keywords. We chose predefined terms highly likely to belong to a toxic tweet in Brazilian Twitter, such as \textit{gay} (\textit{``Gay tem que apanhar'' -- ``Gay should be beaten up''}), \textit{mulherzinha} (\textit{``Mulherzinha, vai lavar louça'' -- ``Sissy, go wash dishes''}), and \textit{nordestino} (\textit{``Nordestino preguiçoso'' -- ``Lazy Northeastern''}). However, using this strategy alone may hinder learning a model capable of generalising the concept of toxicity beyond the scope of keywords. 
Consequently, another strategy was adopted: we scraped tweets that mention influential users like Brazil's president Jair Bolsonaro and soccer player Neymar Jr, prone to receive abuse (around $50$ influential users were monitored). Tweets collected through this method have no restrictions in terms of keywords and should broaden the scope of the data.

We collected more than $10$M unique tweets and randomly selected $21$K examples to compose the annotated corpus. We note that $12,600$ of these posts ($60\%$) comes from the first strategy -- predefined keywords -- and the remaining are tweets from threads of predefined users. The data was pseudoanonymised before being sent for annotation, with all \textit{@} mentions replaced by \textit{@user}.

\subsection{Corpus annotation}
The annotation process started by choosing volunteers to perform the task of assigning labels for each example. For this, we made a public consultation at the Federal University of S\~{a}o Carlos (Brazil) to find candidate annotators ($129$ volunteers registered for the task). From these candidates, $42$ were selected based on their demographic information, aiming to balance annotation bias as the interpretation of toxicity may vary. Each annotator labelled $1,500$ tweets, selecting one of the following categories: \textit{LGBTQ+phobia}, \textit{obscene}, \textit{insult}, \textit{racism}, \textit{misogyny} and/or \textit{xenophobia} (or leaving it blank for \textit{none}). Each tweet was annotated by three different annotators. 

\begin{figure}[ht]
\centering
\includegraphics[width=.4\textwidth]{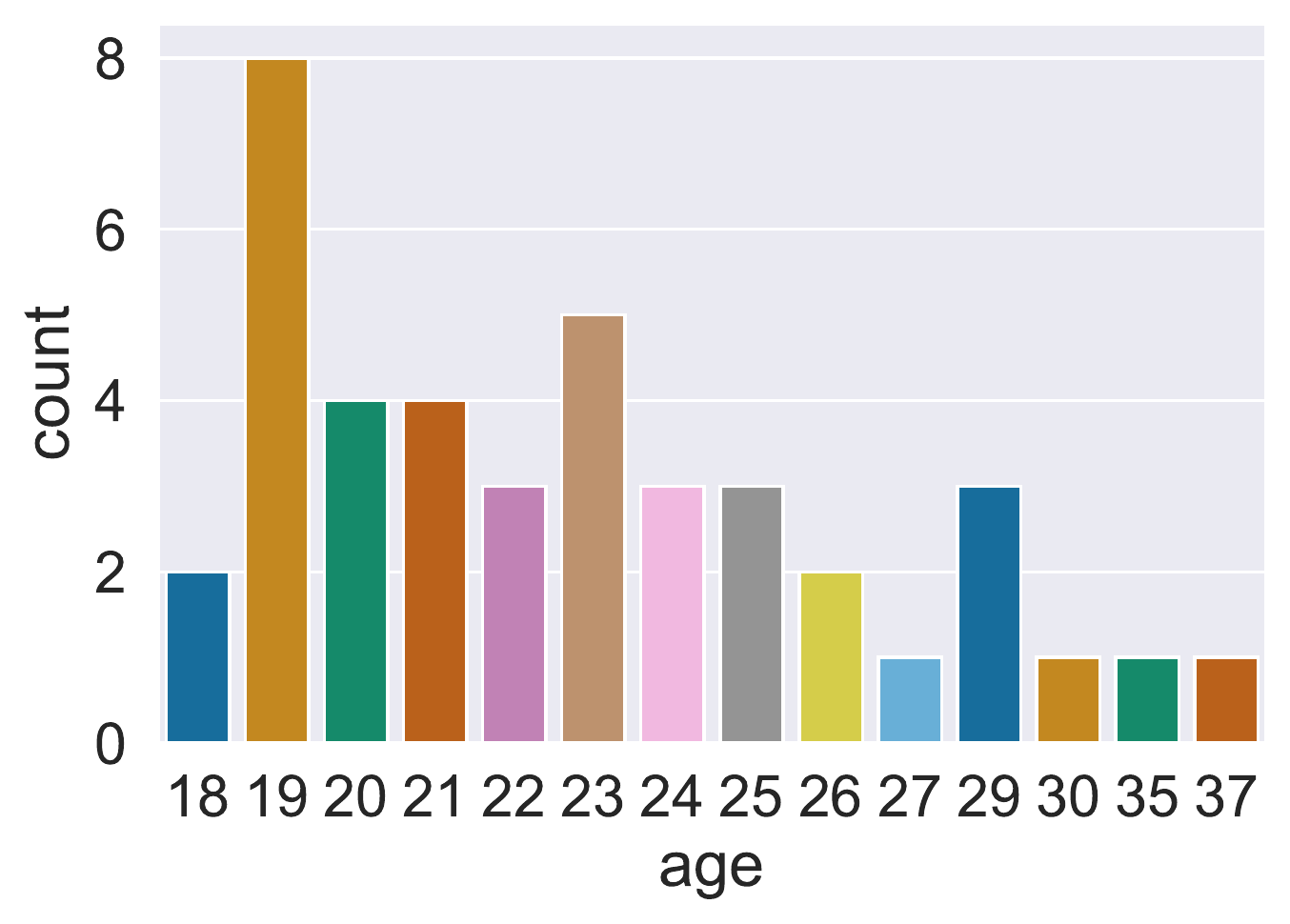}
\caption{Annotators age distribution.}
\label{fig:age}
\end{figure}

To evaluate the diversity among the annotators, we explore their profile. We emphasise that the identity of all annotators has been preserved. At this stage, we only survey general aspects of the volunteers who joined the labelling process.
%After the first call for volunteers, we tried to balance these characteristics among the candidates. Some recorders needed to be replaced due to various factors, causing a slight imbalance in the final set of annotators. For example, as most candidates were white and heterosexual, we attempted a balance between white and non-white, and heterosexual and non-heterosexual people.
Table~\ref{tab:demographics} presents the distribution of annotators regarding sex, sexual orientation, and ethnicity. To define these categories, we use the same values as the Brazilian Institute of Geography and Statistics,\footnote{\url{https://www.ibge.gov.br/en/home-eng.html}} in addition to giving the candidate the option of not declaring a value for each characteristic. Although we tried to keep the demographic aspects as balanced as possible when selecting the annotators, our pool of volunteers was still biased towards people identified as \textit{white} and \textit{heterosexual} (\textit{sex} is a more balanced aspect than the others). The age of the annotators varies between $18$ and $37$ years, with most of them in the range between $19$ and $23$. Figure~\ref{fig:age} illustrates the age distribution.

\begin{table}[ht]
\centering
\setlength{\tabcolsep}{1pt}
\begin{tabular}{p{2cm}lc}
\toprule
& Categories & \# annotators                                                \\ \bottomrule
\multirow{2}{*}{Sex} & Male & 18 \\  
& Female & 24    \\ \midrule

%Sex & Male / Female & 18 / 24                                                      \\ \midrule
& Heterosexual & 22 \\ 
Sexual & Bisexual & 12 \\ 
orientation & Homosexual & 5 \\ 
& Pansexual &  3\\ \midrule

\multirow{5}{*}{Ethnicity}                                                     & White & 25 \\ 
& Brown & 9 \\ 
& Black & 5 \\ 
& Asian & 2\\ 
& Non-Declared & 1  \\ \bottomrule
\end{tabular}
\caption{Annotators demographic information.}\label{tab:demographics}
\end{table}

\begin{table}[!ht]
    \centering
    \begin{tabular}{@{}lr@{}}
    \toprule
                 & $\alpha$ \\ \midrule
    LGBTQ+phobia & 0.68  \\
    Insult       & 0.56  \\
    Xenophobia   & 0.57  \\
    Misogyny     & 0.52  \\
    Obscene      & 0.49  \\
    Racism       & 0.48  \\ \midrule
    Mean         & 0.55  \\ \bottomrule
    \end{tabular}
    \captionof{table}{\textit{Krippendorff}'s $\alpha$ for each label.} \label{tab:kappa}
\end{table}

%Once we defined the dataset, we can explore its main features.
We perform different data analysis over the dataset to better understand its properties. Inter-annotator agreement is calculated in terms of \textit{Krippendorf}’s $\alpha$ (Table~\ref{tab:kappa}), since $\alpha$ is robust to multiple annotators, different degrees of disagreement and, missing values \cite{artstein-poesio-2008-survey}. %presents the $\alpha$ This metric 
%We begin with an analysis of the agreement between the annotators. For this, we calculated Krippendorf’s $\alpha$ for each label, which are shown in . 

%%%%%%%%%%%%%%%%%%%%%%%%%%%%%%
%%%%%%%%%%%%%%%%%%%%%%%%%%%%%%
%%%%%%%%%%%%%%%%%%%%%%%%%%%%%%
\begin{table*}[!ht]
\centering
\setlength{\tabcolsep}{2pt}
\begin{tabular}{@{}p{11.3cm}ccc@{}}
\toprule
           & Ann 1 & Ann 2 & Ann 3 \\ \midrule
\textit{o fdp do filho dela nao parava de tocar auto pra c*****o [...]} & Insult      & None        & Obscene     \\
\footnotesize{\textit{her sob son did not stop to play loud as f**k [...]}} & & & \\
\hline
\textit{[...] VAI SE F***R IRMÃO VC NÃO É FELIZ PQ NAO QUER}  & Obscene     & Insult      & Insult      \\
\footnotesize{\textit{[...] f**k you brother you are not happy because you do not want to be}} & & & \\
\hline
\textit{``Aonde tem um monte que fala mal, mas ninguém vai embora do morro.''}\\ \textit{acha que alguém mora aqui por que quer, c*****o!? Que idéia. [...]} & Obscene     & Obscene     & Insult      \\ 

\footnotesize{\textit{``Where there are loads saying bad things, but nobody leaves the slum.''}} \\ \footnotesize{\textit{who thinks that someone lives here because they want, f**k!? What an idea. [...] }} & & & \\

\bottomrule
\end{tabular}
\captionof{table}{Example of annotation divergence.}\label{tab:divergence}
\end{table*}

\begin{table*}[!ht]
\centering
\begin{tabular}{@{}cccccc@{}}
\toprule
LGBTQ+phobia  & Obscene     & Insult         & Racism       & Misogyny        & Xenophobia      \\ \midrule
viado (59)    & porra (332)    & puta (221)    & nego (6)    & putinha (38)   & sulista (12)   \\
boiola (15)   & caralho (317)   & caralho (150) & branco (6)  & puta (22)      & carioca (7)     \\
viadinho (13) & puta (268)  & cara (135)    & preto (4)   & piranha (19)   & fala (4)        \\
sapatão (12)  & tomar (136) & porra (122)   & nada (4)    & mulher (11)    & paulista (4)   \\
caralho (11)  & fuder (98)   & lixo (101)    & negão (3)   & vagabunda (11) & gente (3)      \\
cara (10)     & cara (94)   & filho (92)    & cara (3)    & quer (8)       & nordestino (3) \\
quer (9)      & merda (90)  & burro (87)    & falando (3) & vaca (8)       & todo (3)       \\
homem (9)     & mano (87)   & tomar (86)    & vida (3)    & fica (6)       & ainda (3)      \\
todo (9)      & toma (85)   & merda (78)    & segue (2)   & onde (5)       & sendo (2)      \\
bicha (9)     & fazer (77)  & idiota (76)   & página (2)  & tudo (5)       & dança (2)      \\ \bottomrule
\end{tabular}
\captionof{table}{The most common words of each class and the number of sentences they occur (within parentheses).} \label{tab:frequency}
\end{table*}
%%%%%%%%%%%%%%%%%%%%%%%%%%%%%%
%%%%%%%%%%%%%%%%%%%%%%%%%%%%%%
%%%%%%%%%%%%%%%%%%%%%%%%%%%%%%

The \textit{LGBTQ+phobia} class shows the highest agreement, which may indicate that comments in this class have a more distinctive lexicon than other classes. The lowest agreement is seem in \textit{obscene} and \textit{racism} classes. Besides, we observed in the annotations many cases in which some examples were labelled as separate classes, although they intend to point the same concept. Classes like \textit{obscene} and \textit{insult} seem to have confused the annotators, which may indicate an intersection in these concepts. Table \ref{tab:divergence} shows examples of disagreements in the classification of \textit{obscene} and \textit{insult}. 

Table~\ref{tab:frequency} presents the ten most frequent words for each class, after removing stopwords. It confirms the intersection between classes \textit{obscene} and \textit{insult}, with six out of ten words in common. For a quantitative analysis, Table \ref{jaccard} presents the \textit{Jaccard} distance between the $100$ most frequent words for each class. \textit{Obscene} and \textit{insult} show a considerably lower distance than other pairs ($0.57$), indicating that they have more words in common.

\begin{table}
\centering
\begin{tabular}{@{}ccccccc@{}}
\toprule
                       & a    & b    & c    & d    & e    & f    \\ \midrule
\multicolumn{1}{c|}{a} & 0.00 & 0.73 & 0.78 & 0.90 & 0.80 & 0.94 \\
\multicolumn{1}{c|}{b} & - & 0.00 & 0.57 & 0.84 & 0.77 & 0.90 \\
\multicolumn{1}{c|}{c} & - & - & 0.00 & 0.86 & 0.75 & 0.92 \\
\multicolumn{1}{c|}{d} & - & - & - & 0.00 & 0.87 & 0.95 \\
\multicolumn{1}{c|}{e} & - & - & - & - & 0.00 & 0.94 \\
%\multicolumn{1}{c|}{f} & - & - & - & - & - & 0.0  \\
\bottomrule
\end{tabular}
\captionof{table}{\textit{Jaccard} distance between all pair of classes. \\
(a) LGBTQ+phobia; (b) Obscene; (c) Insult; (d) Racism; (e) Misogyny; (f) Xenophobia.} \label{jaccard}
\end{table}

\subsection{Dataset characteristics}

\begin{table*}[ht]
\centering
\begin{tabular}{@{}cccccccc@{}}
\toprule
  & LGBTQ+phobia & Insult & Xenophobia & Misogyny & Obscene & \multicolumn{1}{c|}{Racism} & Toxic \\ \midrule
\multicolumn{8}{c}{\textbf{At least one annotator}}                                                       \\ \midrule
0 & 20656      & 16615  & 20849      & 20537    & 14348   & \multicolumn{1}{c|}{20862}  & 11745 \\
1 & 344        & 4385   & 151        & 463      & 6652    & \multicolumn{1}{c|}{138}    & 9255  \\ \midrule
\multicolumn{8}{c}{\textbf{At least two annotators}}                                                      \\ \midrule
0 & 20824      & 19131  & 20958      & 20867    & 18597   & \multicolumn{1}{c|}{20967}  & 16566 \\
1 & 176        & 1869   & 42         & 133      & 2403    & \multicolumn{1}{c|}{33}     & 4424  \\ \midrule
\multicolumn{8}{c}{\textbf{Three annotators}}                                                            \\ \midrule
0 & 20926      & 20483  & 20985      & 20971    & 20388   & \multicolumn{1}{c|}{20994}  & 19510 \\
1 & 74         & 517    & 15         & 29       & 612     & \multicolumn{1}{c|}{6}      & 1490  \\ \bottomrule
\end{tabular}
\captionof{table}{Dataset distribution considering different types of label aggregation.}\label{tab:labels}
\end{table*}

For the purpose of training models for automatically classifying toxic comments, we must create aggregated annotations to provide only one binary label for each class. Different rules can be employed to aggregate the annotations, with different semantics. When we set an example as positive for toxicity only when all the annotators consider it to have the same category of offence, we insert bias to the model to not accuse a comment as toxic unless the offence is evident. Since this is very restrictive, we can also use the majority rule, but there must still be a consensus among the annotators. A last option is to consider that only a positive annotation is sufficient to label the example as positive. This procedure acknowledges that annotators may have divergent views about what was said. It is a risky rule if we intend to create rigid systems that classify the tweets and take corrective or prohibitive actions. However, it is beneficial for training a model that ``raises a flag'' to help moderators to assess the comments. Table~\ref{tab:labels} shows the data distribution for each label and each aggregation strategy.

For the sake of reproducibility and further usage, ToLD-Br is split into default training ($80$\%), development ($10$\%) and test ($10$\%) sets using a stratified strategy. Besides, the corpus is released with all the annotations. Thus, future users of ToLD-Br will be able to use it with all the labels and with varying levels of agreement between the annotators. In this paper, we consider the least restrictive case, where if at least one annotator marked any offence category in an example, the example is positive for toxicity. Likewise, if a tweet was not tagged in any of these categories, it is considered non-toxic. We believe that it is essential that if any person feels uncomfortable with a post, it should be flagged as having a certain degree of toxicity. Therefore, a model built with this data must be able to identify offensive posts, even for a specific group of people.

\section{Materials and Methods} \label{sec:materials}
%João + all

In this section, we describe the techniques, tools, and other materials used in our experimental evaluation. As mentioned before, we restrict our experiments on the dataset labelled as positive when at least one annotator considers the example as toxic. We then investigate the effects of the number of instances in the training data, different algorithms to train a classification model, various scenarios considering single- and multilingual models, and perform an initial experiment with multi-label classification. 

We use Bag-of-Words (BoW) to represent the examples and an AutoML model to build the baseline model (\texttt{BoW+AutoML}). For this, we use the \texttt{auto-sklearn}\footnote{\url{https://automl.github.io/auto-sklearn}} library~\cite{feurer2019auto}. For our BERT-based models, we use the \texttt{simpletransformers}\footnote{\url{github.com/ThilinaRajapakse/simpletransformers}} library, that allows easy training and evaluation. We use default arguments for parameter tuning and define a seed to allow for reproducibility. Two versions of pre-trained BERT language models are applied: Brazilian Portuguese BERT\footnote{\url{huggingface.co/neuralmind/bert-base-portuguese-cased}} \cite{souza2019portuguese}, and Multilingual BERT\footnote{\url{huggingface.co/bert-base-multilingual-cased}} \cite{Wolf2019HuggingFacesTS}.

ToLD-Br is used to fine-tune BERT-based models for our monolingual experiments, with monolingual BERT (\texttt{BR-BERT}) and multilingual BERT (\texttt{M-BERT-BR}). Although \texttt{M-BERT-BR} refers to the multilingual version of BERT, we refer to these two models as ``monolingual models,'' as we trained using the dataset with Brazilian Portuguese sentences alone.

Using the multilingual model, we also carry out experiments in which we add data in English to train the models either through transfer learning or zero-shot learning. For these experiments we use the OLID data, concatenating the training and test splits into a single dataset. For transfer learning, we merged OLID and ToLD-Br to obtain a model with both languages as input, aiming to assess whether extra data in English helps in building better models (\texttt{M-BERT(transfer)}). For zero-shot learning, OLID is used alone at training time, building a model that did not have access to any data in Brazilian Portuguese (\texttt{M-BERT(zero-shot)}). Through these experiments, we can assess the advantages of monolingual models, whether data from another language can directly benefit the classification, and whether a specific monolingual dataset is necessary or not.

We experiment with different sizes of the training set to assess the influence of the volume of data on the classification. For that, we evaluate the results on random subsets of the data. The size of each partition varies in a range between $10$\% and $100$\% adding $10$\% of the data at each iteration. For each step, we repeat the classification three times to minimise the probability of reporting results obtained by chance. Our best model (\texttt{M-BERT-BR}) is used for this experiment (c.f. Section \ref{sec:results}).

Evaluation for binary classification is done in terms of precision, recall and, $F1$-score per class and macro-$F1$. We also analyse the confusion matrices of our systems in order to better visualise the performance of our models in each class, mainly focusing on an analysis of false negatives. 

Although we mainly focus on binary classification, an initial approach for multi-label classification is also presented. We use the adaptation for the multi-label classification scenario available in \texttt{simpletransformers}. In this case, the transformer's output consists of six neurons, each representing one of the labels. These neurons are considered independent in the training and prediction process. Thus, when an output neuron is activated, we set the label represented by this neuron to positive.
Besides, we evaluate the performance of a baseline based on \texttt{BoW+AutoML}, where we train an AutoML model for multilabel classification. Evaluation is done in terms of \textit{Hamming} loss and average precision \cite{tsoumakas2009mining}.

\section{Results and Discussion} \label{sec:results}
%João (plus our help)
This section shows the results of our experiments in classifying toxic comments using ToLD-Br.

\subsection{Binary Classification}

\begin{figure*}[!ht]
\centering
\begin{subfigure}{.4\columnwidth}
  \centering
  % a
  \includegraphics[width=\textwidth]{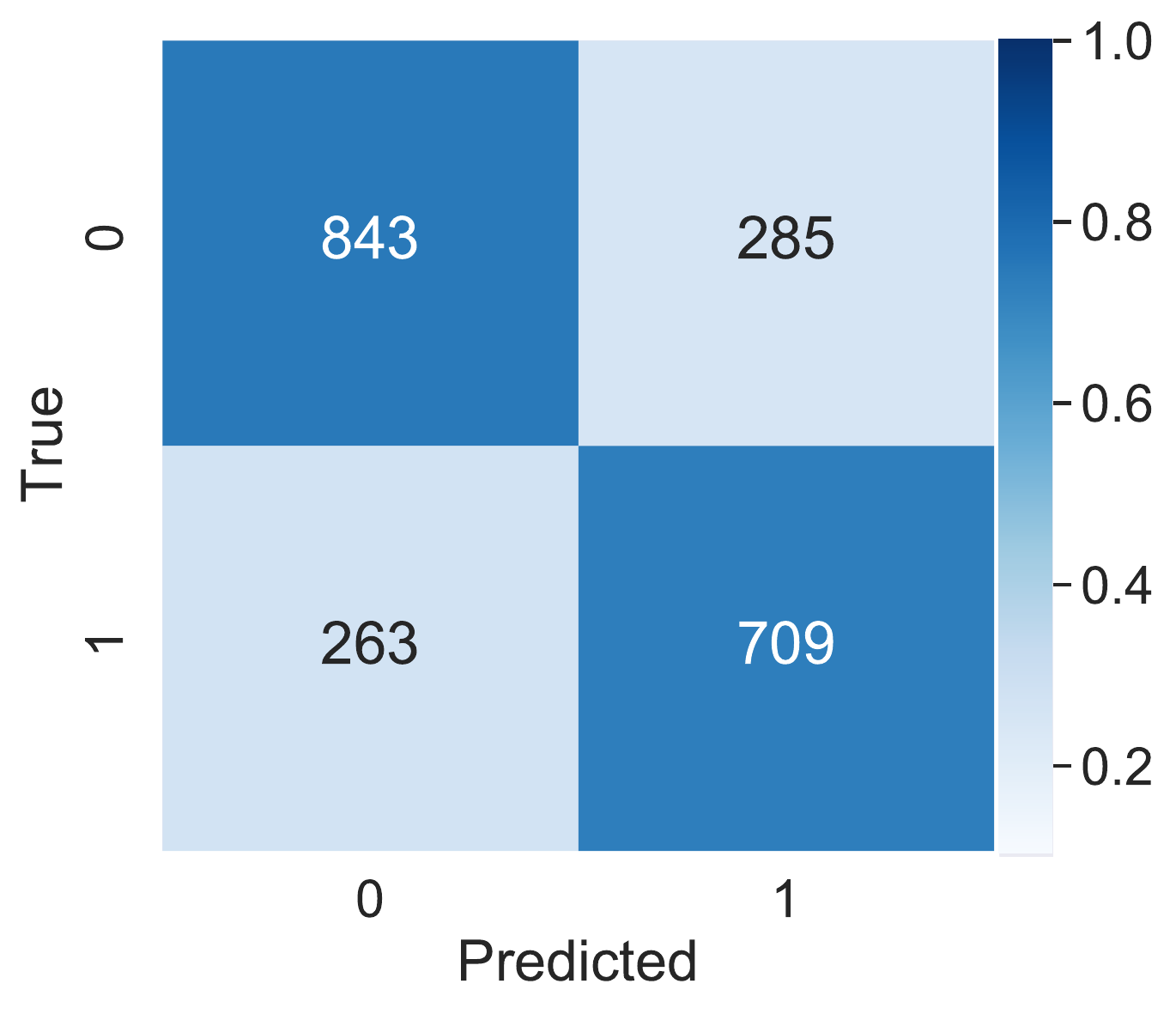}
  \caption{}
  \label{fig:cm-bow}
\end{subfigure}
\begin{subfigure}{.4\columnwidth}
\centering
  % b
  \includegraphics[width=\textwidth]{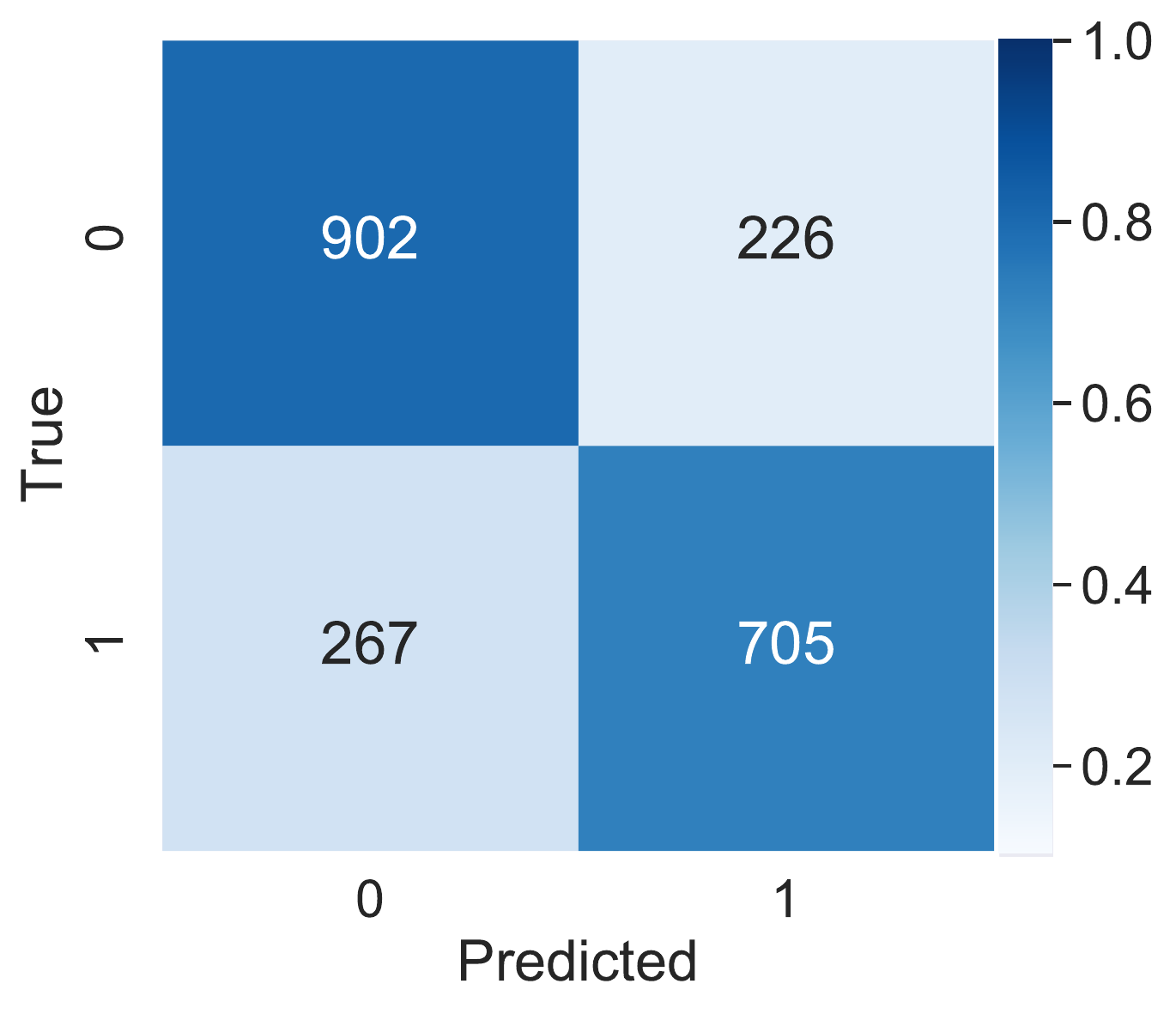}
  \caption{}
  \label{fig:cm-br-bert}
\end{subfigure}
\begin{subfigure}{.4\columnwidth}
  \centering
  % c
  \includegraphics[width=\textwidth]{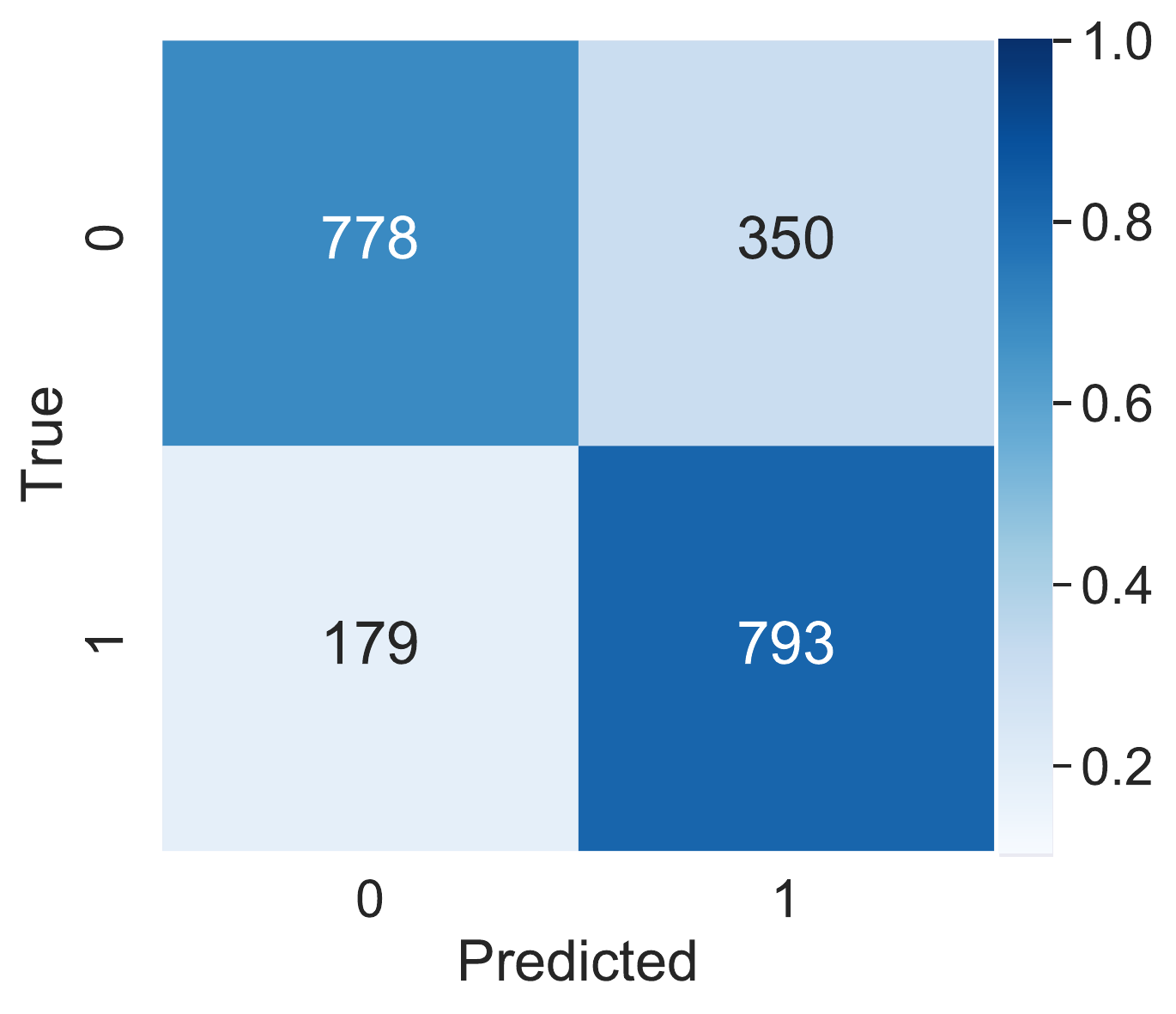}
  \caption{}
  \label{fig:cm-m-bert-br}
\end{subfigure}
\begin{subfigure}{.4\columnwidth}
  \centering
  % d
  \includegraphics[width=\textwidth]{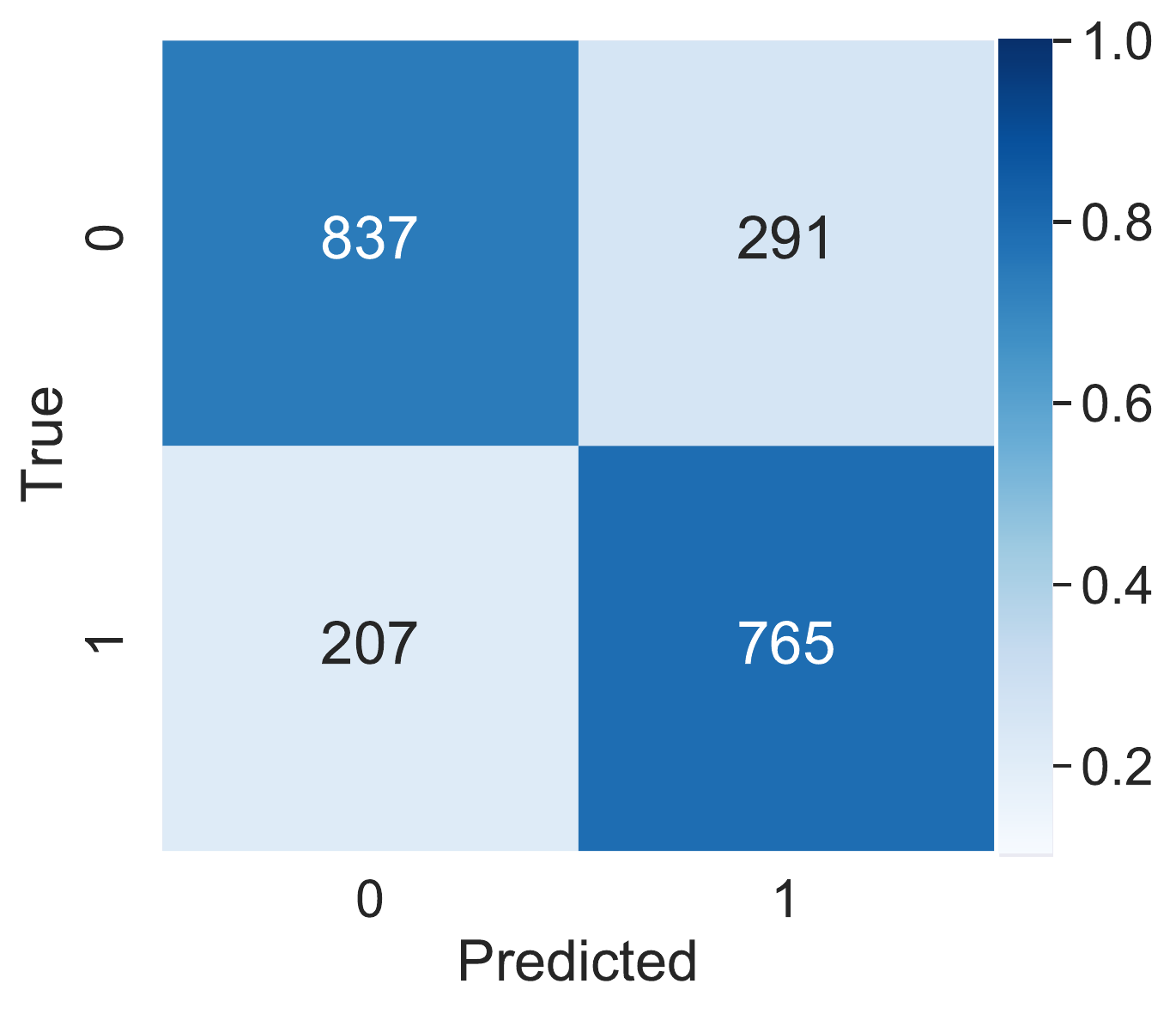}
  \caption{}
  \label{fig:cm-m-bert-zero}
\end{subfigure}
\begin{subfigure}{.4\columnwidth}
  \centering
  % e
  \includegraphics[width=\textwidth]{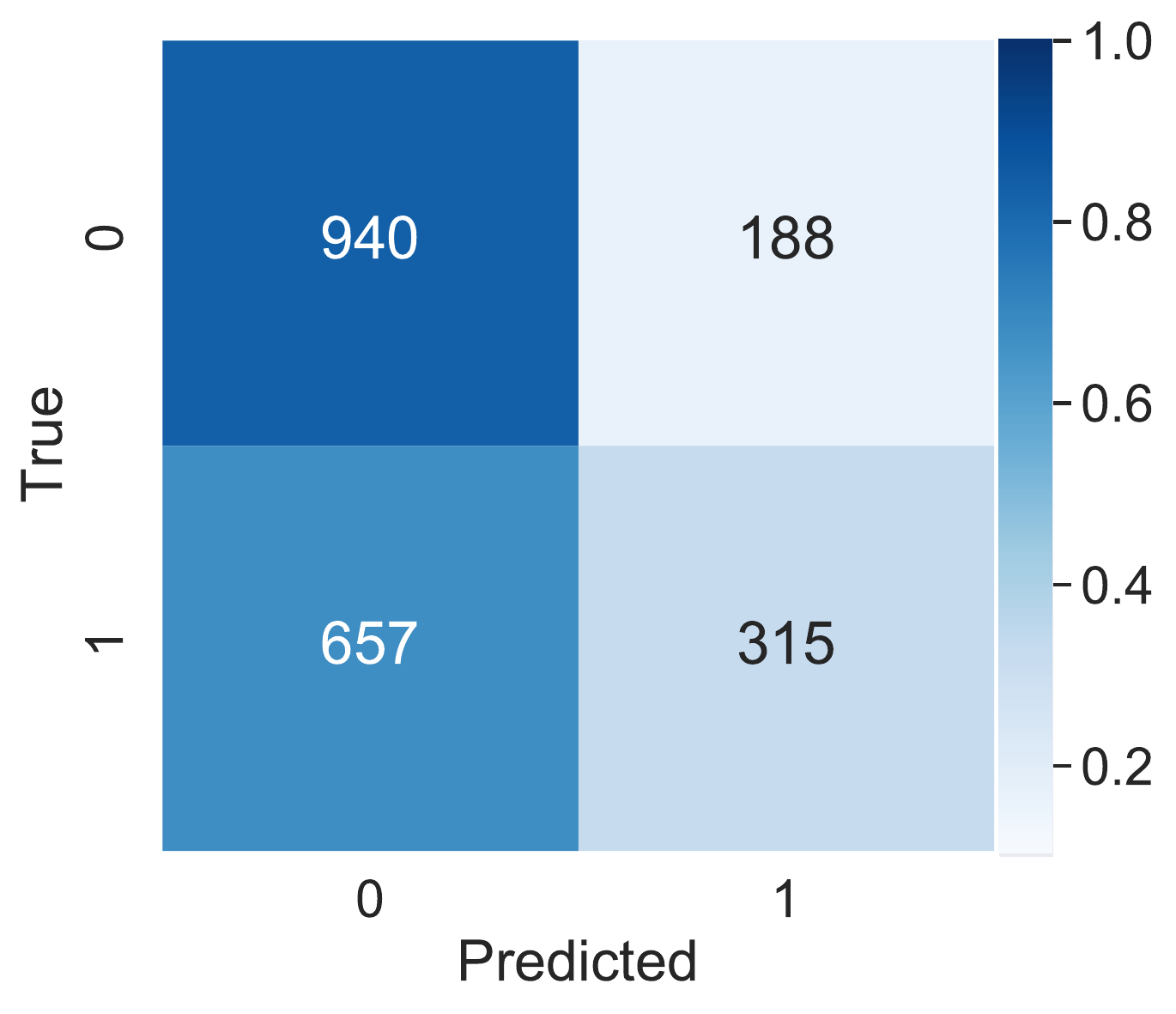} 
  \caption{}
  \label{fig:cm-m-bert-transfer}
\end{subfigure}
\caption{Confusion matrices for each model (a) BoW+AutoML (Baseline); (b) BR-BERT; (c) M-BERT-BR; (d) M-BERT(transfer); (e) M-BERT(zero-shot)}
\label{fig:fig}
\end{figure*}

For evaluating our models, we are particularly interested in models with high performance in the positive class (classification of \textit{toxic} comments). The worst case scenario are false negatives, i.e. \textit{toxic} comments classified as \textit{non-toxic}. Tables~\ref{tab:bow+automl} through~\ref{tab:m-bert-zero} summarises the results for each model. \texttt{BoW+AutoML} is already a competitive model, achieving $74$\% of macro-$F1$, as shown in Table \ref{tab:bow+automl} and Figure \ref{fig:cm-bow}.

\begin{table}[!ht]
\centering
        \begin{tabular}{@{}cccc@{}}\toprule
                     & Precision & Recall & F1-score \\ \midrule
        0            & 0.76      & 0.75   & 0.75     \\
        1            & 0.71      & 0.73   & 0.72     \\ \hline
        Macro Avg    & 0.74      & 0.74   & 0.74     \\
        Weighted Avg & 0.74      & 0.74   & 0.74     \\ \bottomrule
        \end{tabular}
        \captionof{table}{BoW + AutoML}\label{tab:bow+automl}
\end{table}

\begin{table}[!ht]
\centering
        \begin{tabular}{@{}cccc@{}}\toprule
                     & Precision & Recall & F1-score \\ \midrule
        0            & 0.77      & 0.80   & 0.79     \\
        1            & 0.76      & 0.73   & 0.74     \\ \hline
        Macro Avg    & 0.76      & 0.76   & 0.76     \\
        Weighted Avg & 0.76      & 0.77   & 0.76     \\ \bottomrule
        \end{tabular}
        \captionof{table}{BR-BERT}\label{tab:br-bert}
\end{table}

\begin{table}[!ht]
\centering
        \begin{tabular}{@{}cccc@{}}\toprule
                     & Precision & Recall & F1-score \\ \midrule
        0            & 0.81      & 0.69   & 0.75     \\
        1            & 0.69      & 0.82   & 0.75     \\ \hline
        Macro Avg    & 0.75      & 0.75   & 0.75     \\
        Weighted Avg & 0.76      & 0.75   & 0.75     \\ \bottomrule
        \end{tabular}
    \captionof{table}{M-BERT-BR}\label{tab:m-bert-br} 
\end{table}

\begin{table}[!ht]
\centering
        \begin{tabular}{@{}cccc@{}}\toprule
                         & Precision & Recall & F1-score \\ \midrule
            0            & 0.80      & 0.74   & 0.77     \\ 
            1            & 0.72      & 0.79   & 0.75     \\ \hline
            Macro Avg    & 0.76      & 0.76   & 0.76     \\ 
            Weighted Avg & 0.77      & 0.76   & 0.76     \\ \bottomrule
        \end{tabular}
        \captionof{table}{M-BERT(transfer)}\label{tab:m-bert-transfer}
\end{table}

\begin{table}[!ht]
\centering
    \begin{tabular}{@{}cccc@{}}\toprule
                     & Precision & Recall & F1-score \\ \midrule
        0            & 0.59      & 0.83   & 0.69     \\
        1            & 0.63      & 0.32   & 0.43     \\ \hline
        Macro Avg    & 0.61      & 0.58   & 0.56     \\
        Weighted Avg & 0.61      & 0.60   & 0.57     \\ \bottomrule
        \end{tabular}
    \captionof{table}{M-BERT(zero-shot)}\label{tab:m-bert-zero}
\end{table}

\begin{figure*}
\begin{subfigure}{.5\textwidth}
  \centering
  % positive
  \includegraphics[width=0.95\textwidth]{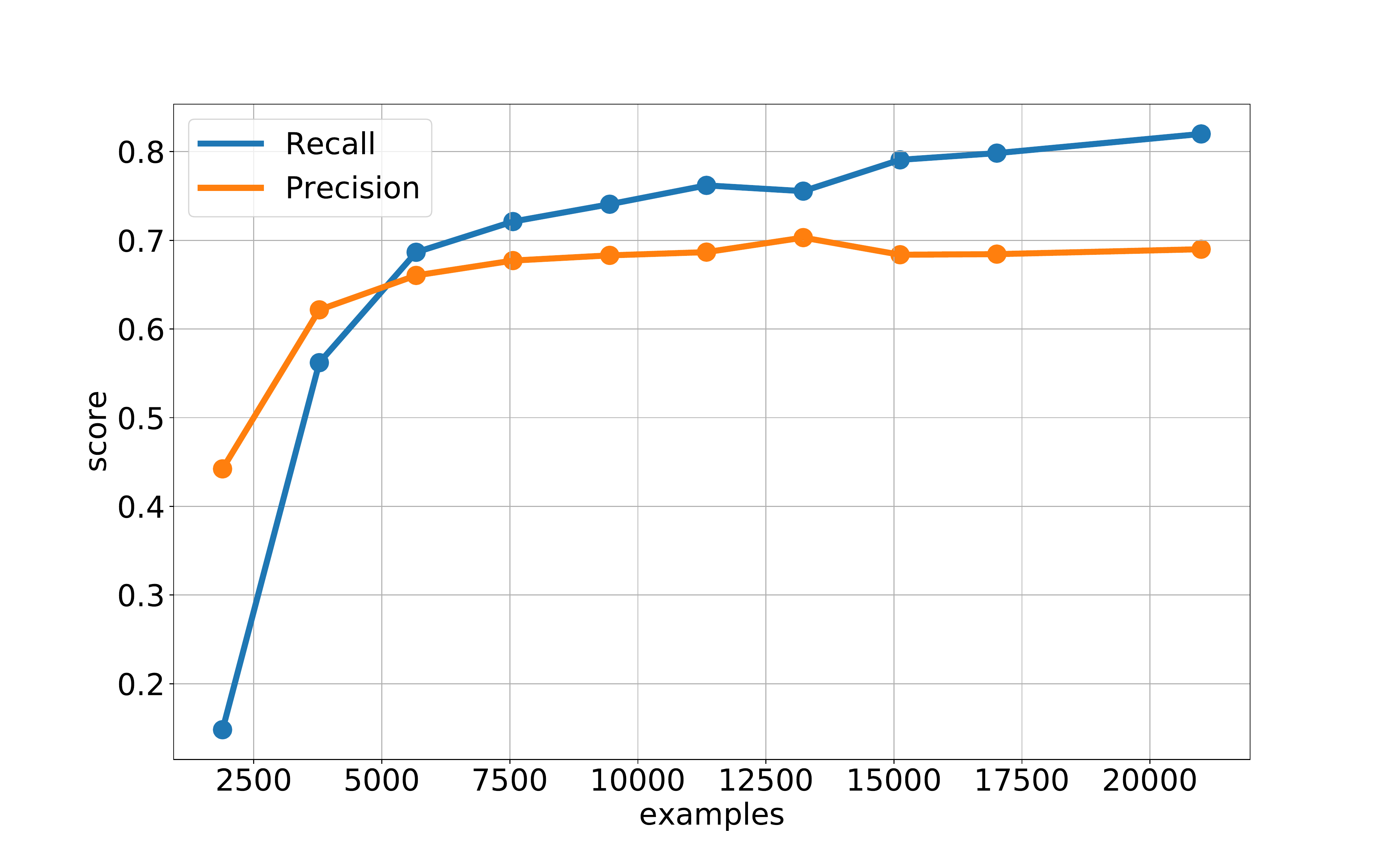}
  \caption{}
  \label{fig:learning-curve-positive}
\end{subfigure}
\begin{subfigure}{.5\textwidth}
  \centering
  % negative
  \includegraphics[width=0.95\textwidth]{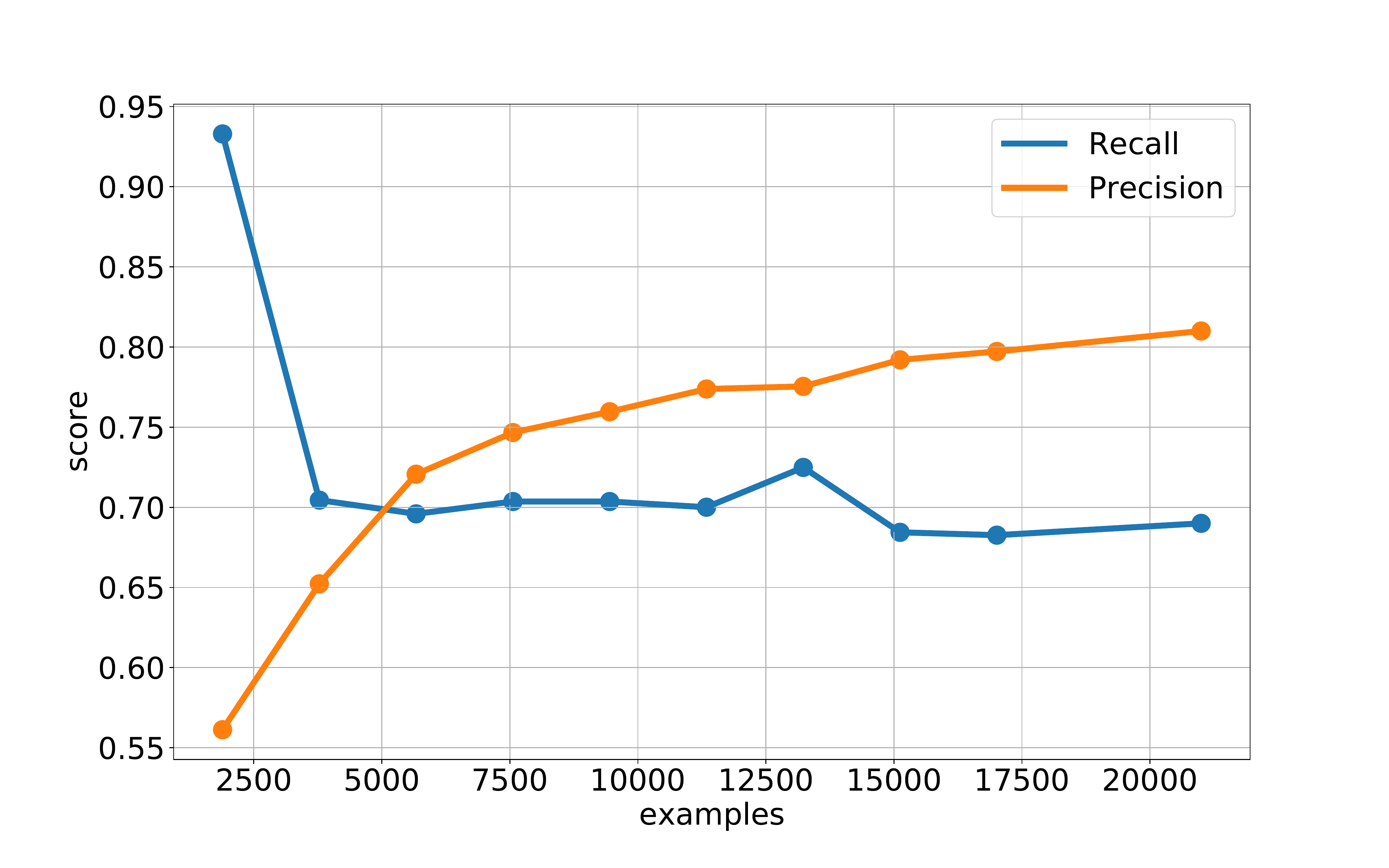}
  \caption{}
  \label{fig:learning-curve-negative}
  \end{subfigure}
\caption{Precision and recall for different sizes of the training dataset for the (a) \textit{positive} and (b) \textit{negative} classes.}
\label{fig:learning-curve}
\end{figure*}

The monolingual models \texttt{BR-BERT} and \texttt{M-BERT-BR} (Tables \ref{tab:br-bert} and \ref{tab:m-bert-br}, respectively) show very similar performances in all metrics, with \texttt{BR-BERT} being slightly better in terms of macro-$F1$. However, \texttt{M-BERT-BR} is better in terms of $F1$-score for the positive class and shows fewer false negatives than \texttt{BR-BERT} (Figure \ref{fig:cm-br-bert} for \texttt{BR-BERT} and Figure \ref{fig:cm-m-bert-br} for \texttt{M-BERT-BR}). 

\texttt{M-BERT(transfer)} (Table \ref{tab:m-bert-transfer}) does not outperform the monolingual models and it also shows more false negatives than \texttt{M-BERT-BR} (Figure \ref{fig:cm-m-bert-transfer}). On the other hand, the number of false negatives in \texttt{BR-BERT} ($267$) is slightly higher than the number of false negatives in \texttt{M-BERT(transfer)} ($207$). Finally, \texttt{M-BERT(zero-shot)} (Table \ref{tab:m-bert-zero}) is the worst model, as expected. It performs particularly bad when classifying the positive class, achieving only $43$\% of $F1$-score for this class, mainly caused by its high number of false negatives (Figure \ref{fig:cm-m-bert-zero}).% ($657$). 

In summary, transfer learning does not seem to improve over the overall performance of monolingual models. Based on the analysis of false negatives, \texttt{M-BERT-BR} appears as our best model. Zero-shot learning shows a very low performance, being particularly bad in the positive class.

\paragraph{Error Analysis}
We also analyse the performance of our best model (\texttt{M-BERT-BR}) in each fine-grained class. The idea is to identify which toxic classes are most difficult to be classified as \textit{toxic} by our binary classifier. As false negatives are a critical type of error in our application, Table \ref{tab:error404} shows the false negative rate (false negatives / expected positives) for each toxic class. %The model's performance for the binary classification is highly influenced by the number of examples for a specific label. 
The ratio of false negatives is inversely proportional to the number of examples for a specific class. \textit{Insult} and \textit{obscene}, the largest classes, show the lowest false negative rate, whilst the highest rates are shown by classes with less examples (\textit{racism} and \textit{xenophobia}). Therefore, in order to improve classification models, these aspects of the imbalanced data need to be taken into account and further studied.  

\begin{table}[!ht]
\centering
\begin{tabular}{@{}lc@{}}
\toprule
             & False negative rate \\ \midrule
LGBTQ+phobia & 7/35 (0.2)                                   \\
Insult       & 67/448 (0.15)                                 \\
Xenophobia   & 13/19 (0.68)                                  \\
Misogyny     & 7/45 (0.15)                                   \\
Obscene      & 117/701 (0.17)                                \\
Racism       & 8/17 (0.47)                                   \\ \bottomrule
\end{tabular}
\captionof{table}{Error analysis for each label.}\label{tab:error404}
\end{table}

\subsection{Importance of Large Datasets}
In this experiment, we highlight the importance of collecting a considerable amount of examples, as toxicity can be expressed in many different ways. We separated the training data into $10$ random splits from $10$\% to $100$\% of the data, increasing $10$\% of data at each step, and trained \texttt{M-BERT-BR} with three random samples for each step. Figure~\ref{fig:learning-curve} shows the mean recall, precision and $F1$-score for the positive and negative classes, respectively, for each data split. With few training examples, the model only performs well on the majority class, but as the number of instances grows, recall for the negative class starts decreasing while recall for the positive class increases, and precision rises for both classes. At least $6$K examples seems to be necessary to achieve reliable results, while previous work for Portuguese reports the largest dataset with only $5,668$ examples. This highlights the importance of ToLD-Br, as a large-scale dataset.

\subsection{Multi-Label Classification}
We experiment with multi-label classification, building a model using the Multilingual BERT (similar to \texttt{M-BERT-BR}). Our baseline is a set of \texttt{BoW+AutoML} models trained using Binary Relevance~\cite{tsoumakas2009mining} for multi-label classification. The BERT-based models adopt a score threshold of $0.5$ in the output neuron to deal with multi-label. If the activation for a label in the output layer is higher than the threshold, we consider it positive.

\begin{figure}[!ht]
\centering

\begin{subfigure}{.4\columnwidth}
  \centering
  % a
  \includegraphics[width=\textwidth]{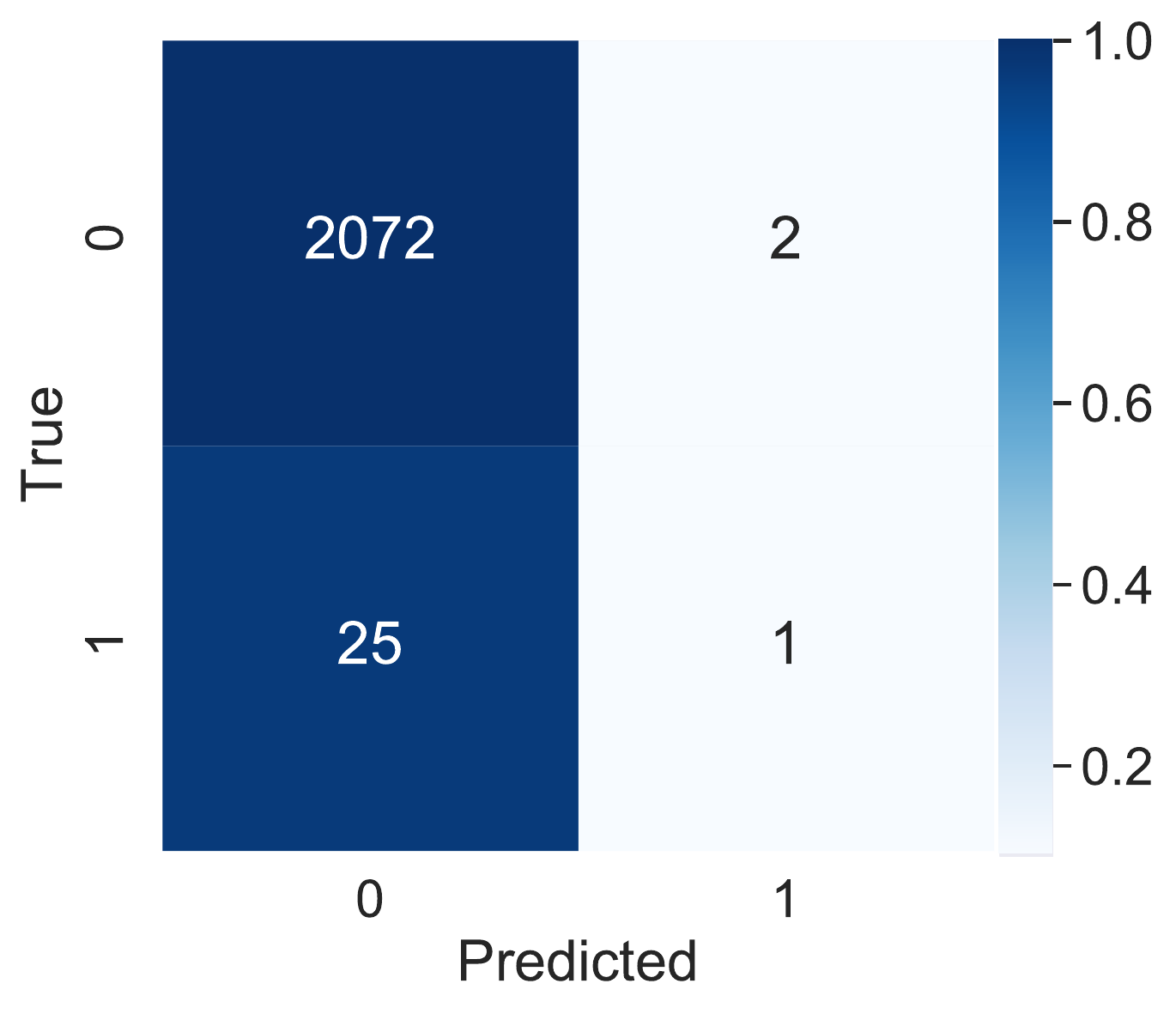}
  \caption{}
  \label{fig:multiclass-homophobia}
\end{subfigure}
\begin{subfigure}{.4\columnwidth}
  \centering
  % b
  \includegraphics[width=\textwidth]{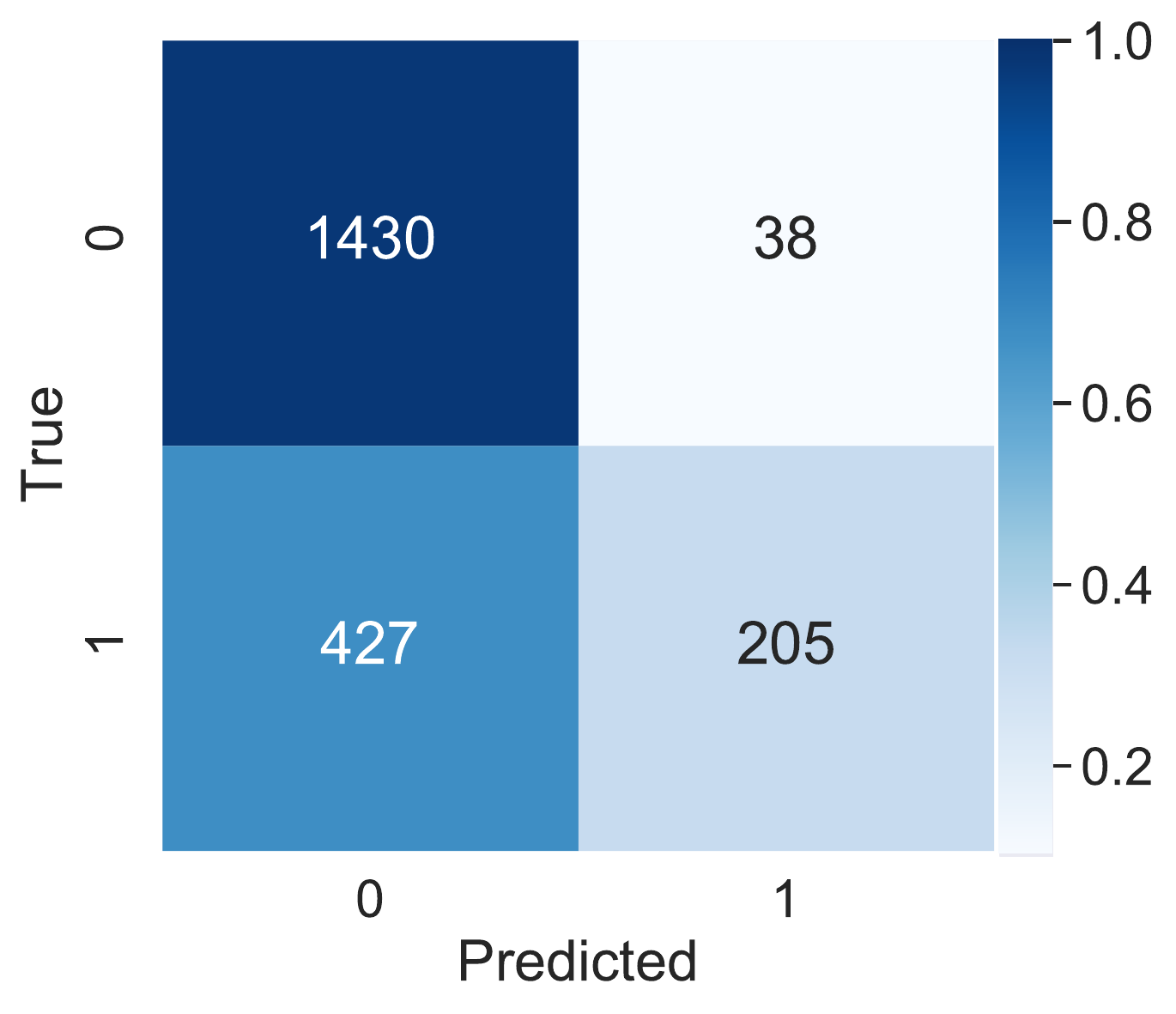} 
  \caption{}
  \label{fig:multiclass-obscene}
\end{subfigure}
\begin{subfigure}{.4\columnwidth}
  \centering
  % c
  \includegraphics[width=\textwidth]{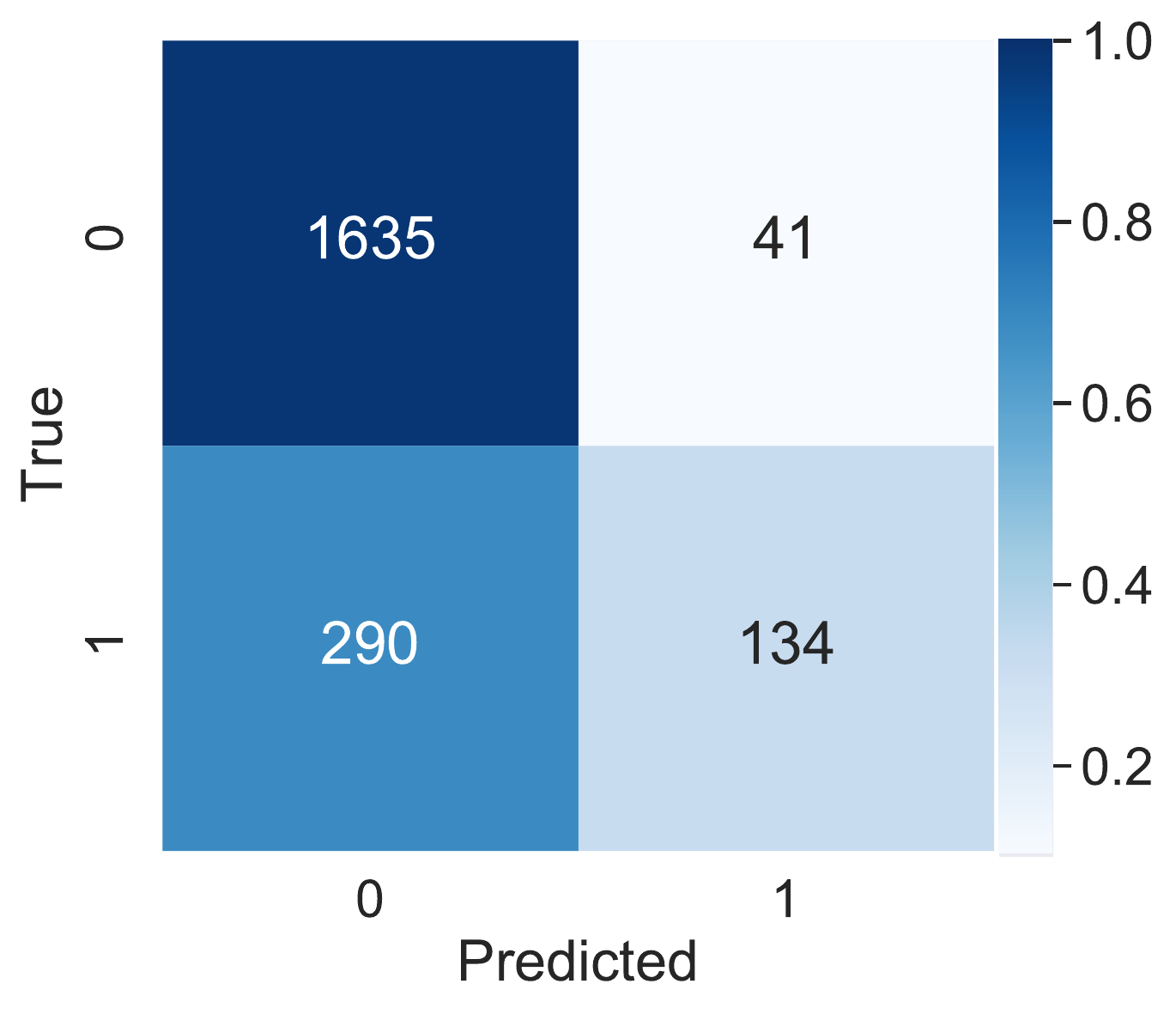}
  \caption{}
  \label{fig:multiclass-insult}
\end{subfigure}
\begin{subfigure}{.4\columnwidth}
\centering
  % d
  \includegraphics[width=\textwidth]{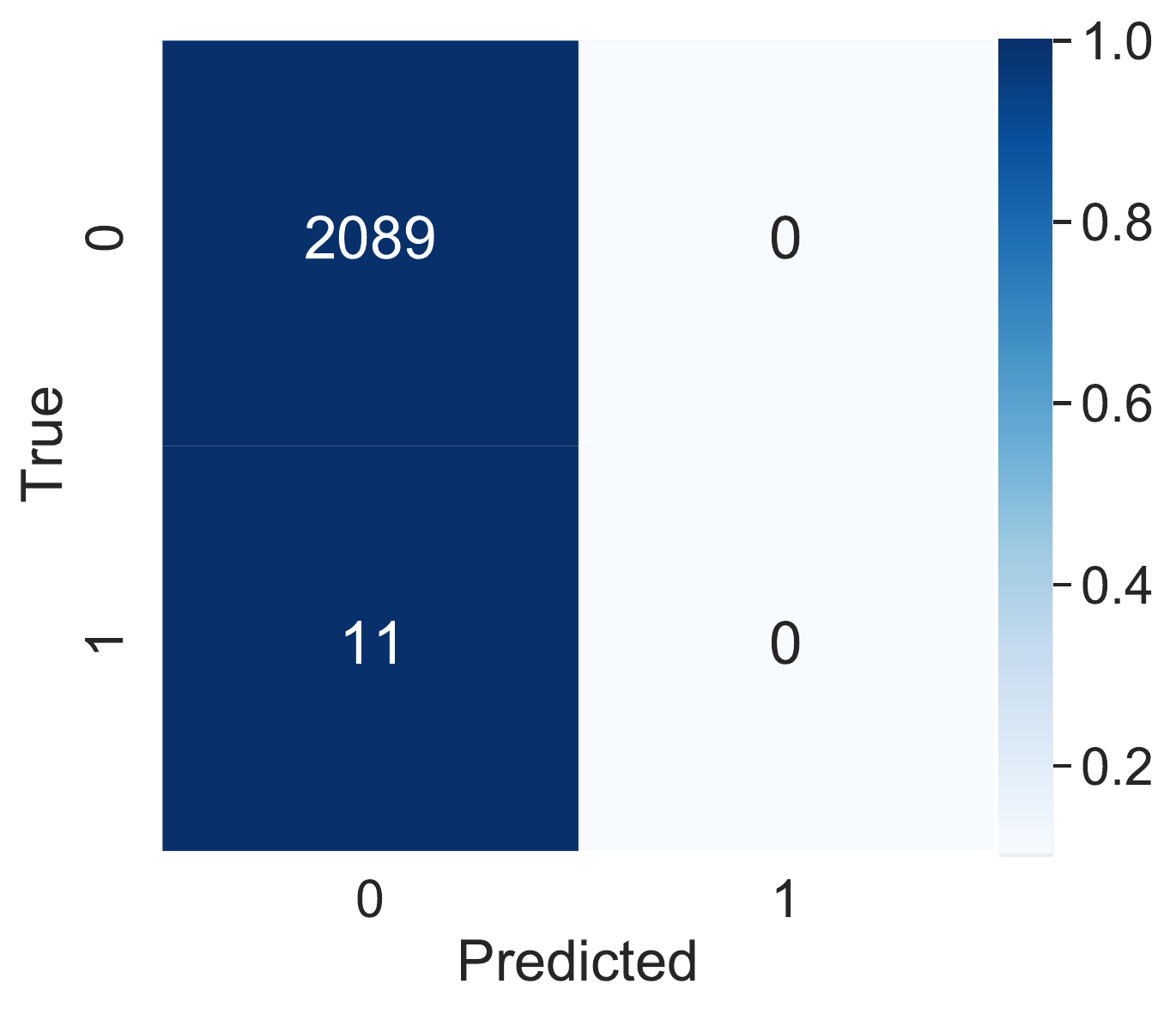}
  \caption{}
  \label{fig:multiclass-racism}
\end{subfigure}
\begin{subfigure}{.4\columnwidth}
  \centering
  % e
  \includegraphics[width=\textwidth]{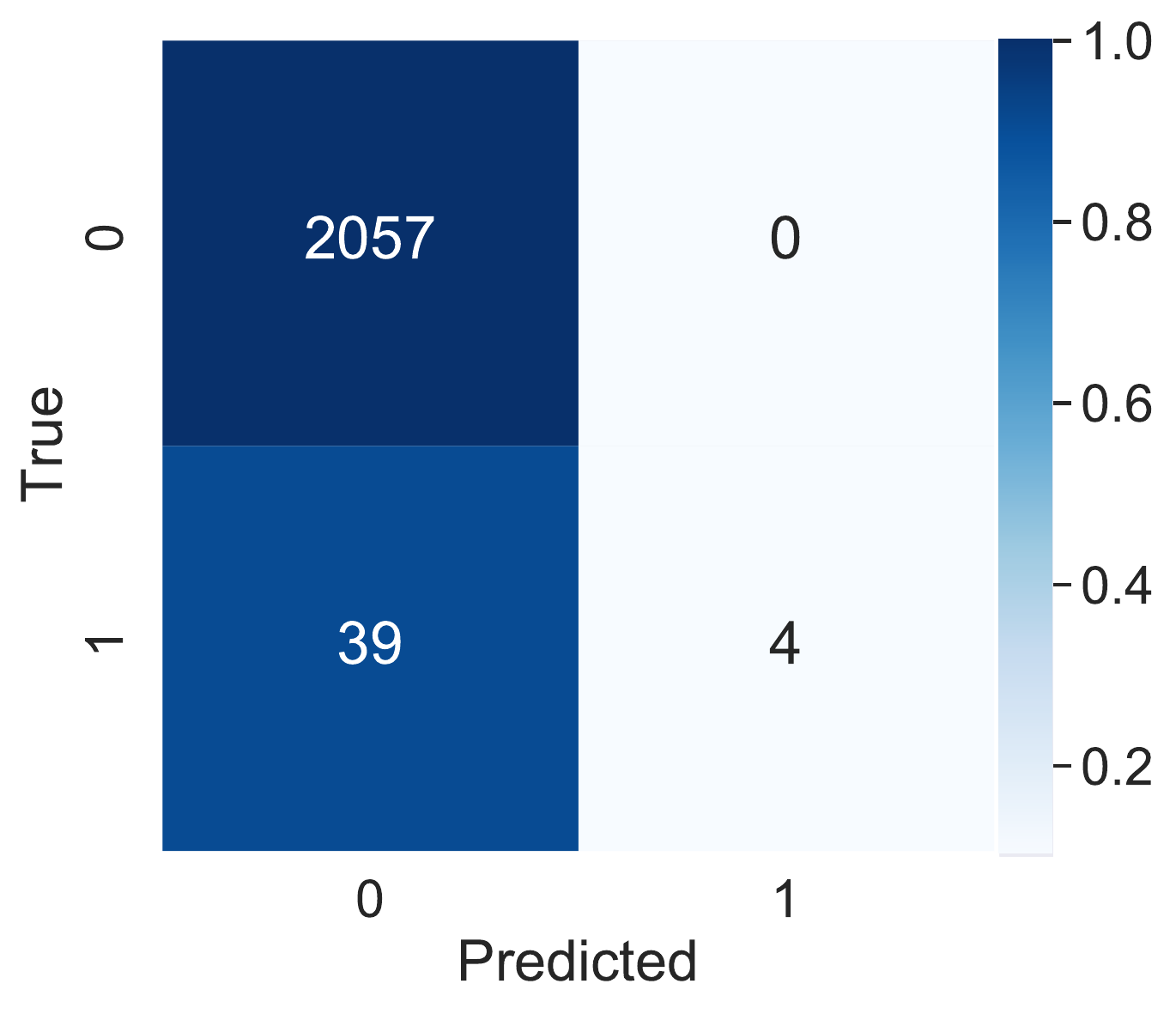}
  \caption{}
  \label{fig:multiclass-misogyny}
\end{subfigure}
\begin{subfigure}{.4\columnwidth}
  \centering
  % f
  \includegraphics[width=\textwidth]{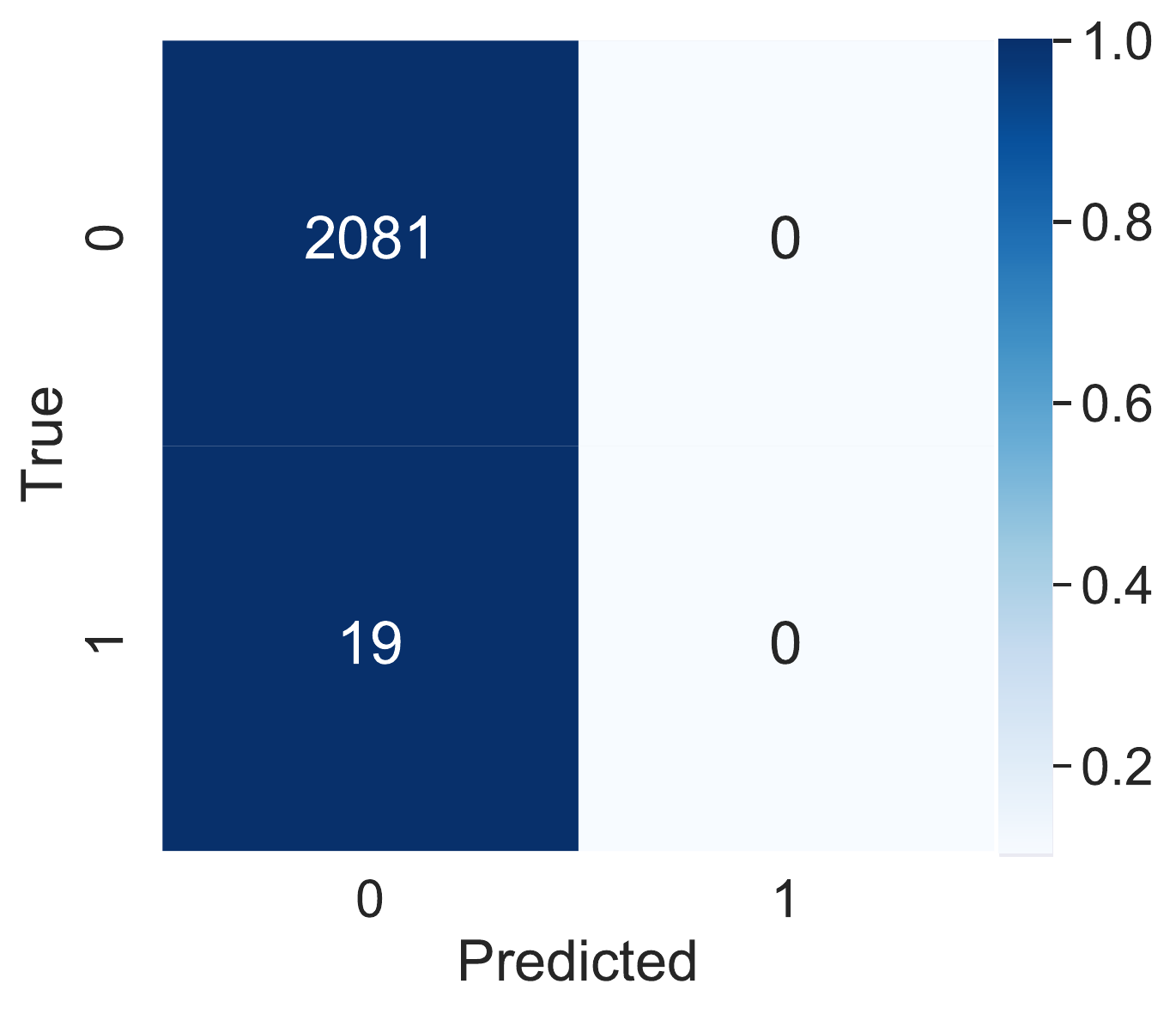}
  \caption{}
  \label{fig:multiclass-xenophobia}
\end{subfigure}
\caption{Confusion matrices for each label (a) \textit{LGBTQ+phobia}; (b) Obscene; (c) Insult; (d) Racism; (e) Misogyny; (f) Xenophobia.}
\label{fig:multilabelcm}
\end{figure}

The baseline model obtained $0.08$ and $0.20$ of \textit{Hamming} loss and average precision, respectively, while \texttt{M-BERT-BR} resulted in $0.07$ and $0.19$ for these measures, respectively. Figure~\ref{fig:multilabelcm} displays the confusion matrices obtained by \texttt{M-BERT-BR}.

This scenario is considerably more challenging than binary classification. The positive class of each label corresponds to a subset of the examples labelled as \textit{toxic}. Thus, it is likely that the number of instances for these classes will be insufficient for the model to learn. Besides, the problem of unbalanced classes becomes evident (c.f. Table~\ref{tab:labels}). As a consequence, it is clear that labels with a small number of positive examples, like \textit{racism}, \textit{misogyny}, \textit{xenophobia}, and \textit{LGBTQ+phobia} were almost entirely classified as negative. In contrast, for \textit{obscene} and \textit{insult}, labels with a considerable amount of positive examples, the model was capable of classifying some examples correctly. In all cases, besides \textit{insult}, the baseline performs slightly better for the positive class (which justify the higher \textit{Hamming} loss). This setback is likely due to the difficulty of the neural model to learn with few examples.

\section{Concluding Remarks}\label{sec:conclusion}
In this paper, we present ToLD-Br: a dataset for the classification of toxic comments on Twitter in Brazilian Portuguese. Through a wide and comprehensive analysis, we demonstrated the need for this dataset for studies on automatic classification of toxic comments. We highlight that monolingual approaches for this task still outperform multilingual experiments and that large-scale datasets are needed for building reliable models. Also, we show that there are still challenges to be overcome, such as the naturally significant class imbalance when dealing with multi-label classification.

As future work, in addition to deal with class imbalance, we intend to evaluate if aggregating classes with high divergences between annotators can build more reliable models. Besides, we intend to assess the benefits of adding unlabelled data to ToLD-Br to use semi-supervised techniques. %Finally, we will carry out a broader experimental analysis, with different models to seek better results in this application. Our goal is to build knowledge and provide resources to improve tools to fight the toxicity in social media.

\section{Acknowledgements}\label{sec:acknowledgements}
We thank the volunteers from UFSCar that made this research possible. The MIDAS group\footnote{\url{midas.ufscar.br}} from the Federal University of São Carlos (UFSCar), Brazil, funded the annotation process. The SoBigData TransNational Access program (EU H2020, grant agreement: 654024) funded Diego Silva and Jo\~{a}o Leite's visits to the University of Sheffield.
%We would like to thank SoBigData (http://project.sobigdata.eu/) and the University of Sheffield's Computer Science Department for funding and hosting the author, providing access to GATE Cloud’s Twitter Collector as well as the infrastructure and personnel, specially from the Natural Language Processing research group that kindly helped us through this research. We also thank Universidade Federal de São Carlos and the volunteers that made this research possible. Finally, we express our gratitude towards the machine learning research group MIDAS (midas.ufscar.br) at Universidade Federal de São Carlos for funding the annotation process.

\bibliography{anthology,aacl-ijcnlp2020}
\bibliographystyle{acl_natbib}

\end{document}